\documentclass[runningheads]{llncs}
\usepackage[T1]{fontenc}
\usepackage{graphicx}
\renewcommand{\bfseries}{\scshape}

\usepackage{textcomp}
\usepackage{ccfonts} 
\usepackage{concmath} 
\usepackage{amsmath}

\usepackage{booktabs}
\usepackage{amssymb}

\usepackage{hyperref} 

\usepackage[table]{xcolor}   
\usepackage{tabularx}       
\usepackage{booktabs}

\begin{document}
\title{Neighbor-Aware Informal Settlement Mapping with Graph Convolutional Networks}
\titlerunning{Neighbor-Aware Informal Settlement Mapping with GCNs}
%
\author{
Thomas Hallopeau\inst{1}\orcidID{0000-0000-0000-0000} \and
Joris Guérin\inst{1}\orcidID{0000-0002-8048-8960} \and
Laurent Demagistri\inst{1}\orcidID{0000-0003-1910-200X} \and
Christovam Barcellos\inst{2}\orcidID{0000-0002-1161-2753} \and
Nadine Dessay\inst{1}\orcidID{0000-0003-0526-3531}
}
\authorrunning{T. Hallopeau et al.}
%
\institute{
ESPACE-DEV, French National Research Institute for Sustainable Development~(IRD), University of Montpellier, France\\
\email{\{firstname.lastname\}@ird.fr}
\and
ICICT, Oswaldo Cruz Foundation (Fiocruz), Rio de Janeiro, Brazil
}
\maketitle              
\begin{abstract}
Mapping informal settlements is crucial for addressing challenges related to urban planning, public health, and infrastructure in rapidly growing cities. Geospatial machine learning has emerged as a key tool for detecting and mapping these areas from remote sensing data. However, existing approaches often treat spatial units independently, neglecting the relational structure of the urban fabric. We propose a graph-based framework that explicitly incorporates local geographical context into the classification process. Each spatial unit (cell) is embedded in a graph structure along with its adjacent neighbors, and a lightweight Graph Convolutional Network (GCN) is trained to classify whether the central cell belongs to an informal settlement. Experiments are conducted on a case study in Rio de Janeiro using spatial cross-validation across five distinct zones, ensuring robustness and generalizability across heterogeneous urban landscapes. Our method outperforms standard baselines, improving Kappa coefficient by 17 points over individual cell classification. We also show that graph-based modeling surpasses simple feature concatenation of neighboring cells, demonstrating the benefit of encoding spatial structure for urban scene understanding.

\end{abstract}
\section{Introduction}

Informal settlements are rapidly growing urban areas characterized by unregulated construction, high population density, and limited access to infrastructure and services. Mapping these settlements is critical for urban planning, disaster risk management, and the provision of basic services, yet remains a significant technical challenge due to their morphological complexity. The illegal nature of those spontaneous settlements make them difficult to map with field surveys or socio-economic data, and the semantic uncertainty surrounding their definition poses additional difficulties~\cite{gevaertChallengesMappingMissing2019}. 

Recent advances in geospatial machine learning have made it possible to classify informal settlements from spatial data, including satellite imagery~\cite{kufferSlumsSpace152016,rajDeepLearningSlum2024}. Supervised machine learning models using multispectral satellite data and auxiliary data sources have shown promising results in delineating informal areas~\cite{owenApproachDifferentiateInformal2013,hallopeauAddressingDataImbalance2025,rajDeepLearningSlum2024}. However, most existing machine learning methods operate under a grid-based or patch-based paradigm, treating spatial units independently. This often fails to capture the spatial relationships and urban continuity that define the fabric of informal areas.
 
Meanwhile, Graph Neural Networks (GNNs) have emerged as powerful tools for learning on structured data by modeling entities as nodes and their interactions as edges~\cite{zhouGraphNeuralNetworks2020,wuComprehensiveSurveyGraph2021}. In the spatial domain, GNNs offer the potential to encode topological and contextual dependencies between spatial units. However, their application to urban mapping remains limited. While some exploratory works have used GNNs to characterize building arrangements~\cite{yanGraphConvolutionalNeural2019,zhaoRecognitionBuildingGroup2020} or classify building functions~\cite{chenInterpretingCoreForms2025}, these approaches rely on detailed building geometries rarely available in informal settlements and difficult to extract from satellite imagery due to the high building density. Road network–based graph methods have also been explored~\cite{xueQuantifyingSpatialHomogeneity2022,maGraphConvolutionalNetworks2024}, but are mostly limited to analyzing global urban form rather than performing complex and fined-grained intra-urban classification tasks like informal settlement mapping. 



\begin{figure}[t]
\centering
\includegraphics[width=\textwidth]{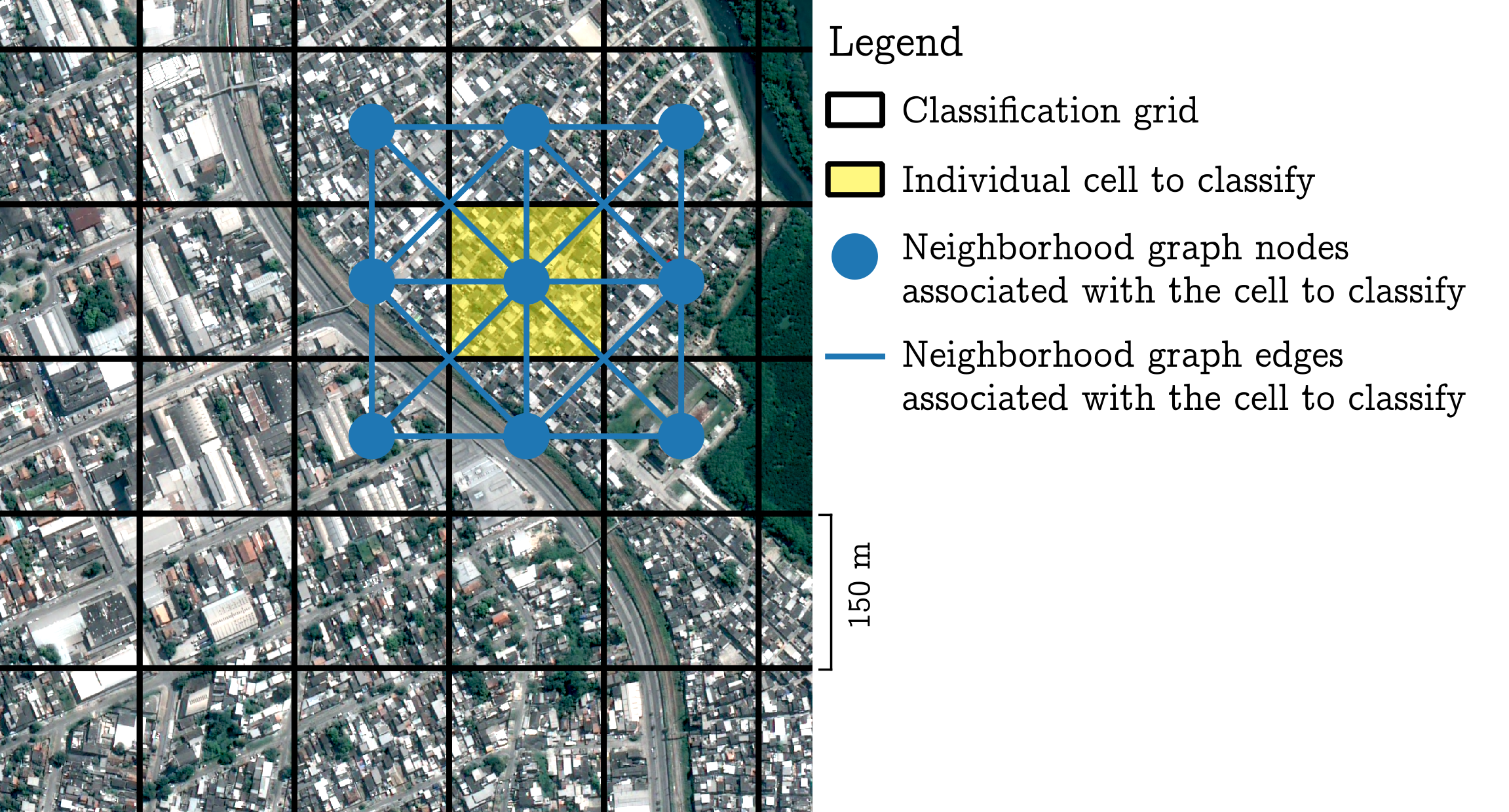}
\caption{Grid-Based Partitioning of the Urban Space and an Example of Graph Constructed around a Specific Cell to Classify} 
\label{fig:GTG}
\end{figure}

In this paper, we explore the use of GNNs for informal settlement mapping by modeling local spatial contexts as structured graphs. We propose a novel method based on graph classification, where each spatial unit (cell) and its adjacent neighbors form a localized graph composed of 9 nodes and 20 edges (see Fig.~\ref{fig:GTG}). Our framework uses Sentinel-2 multispectral data, the Copernicus Digital Elevation Model, and OpenStreetMap road network to compute descriptive features per cell in a $150m$ resolution grid. Each cell is then embedded within a local $3\times3$ graph, and classified using a lightweight Graph Convolutional Network (GCN) designed for this task.

We evaluate our approach on a use case in Rio de Janeiro, a city with highly fragmented and heterogeneous urban morphology. Using a spatial cross-validation methodology across five urban zones with distinct characteristics, we show that including neighborhood cell context by simple feature concatenation improves informal settlement classification compared to individual cell classification. We demonstrate that incorporating spatial structure with a graph-based approach outperforms individual cell classification and feature concatenation of neighboring cells, with improvements in Kappa coefficient by 17 and 10 points.


\section{Method}

\begin{figure}[t]
\centering
\includegraphics[width=\textwidth]{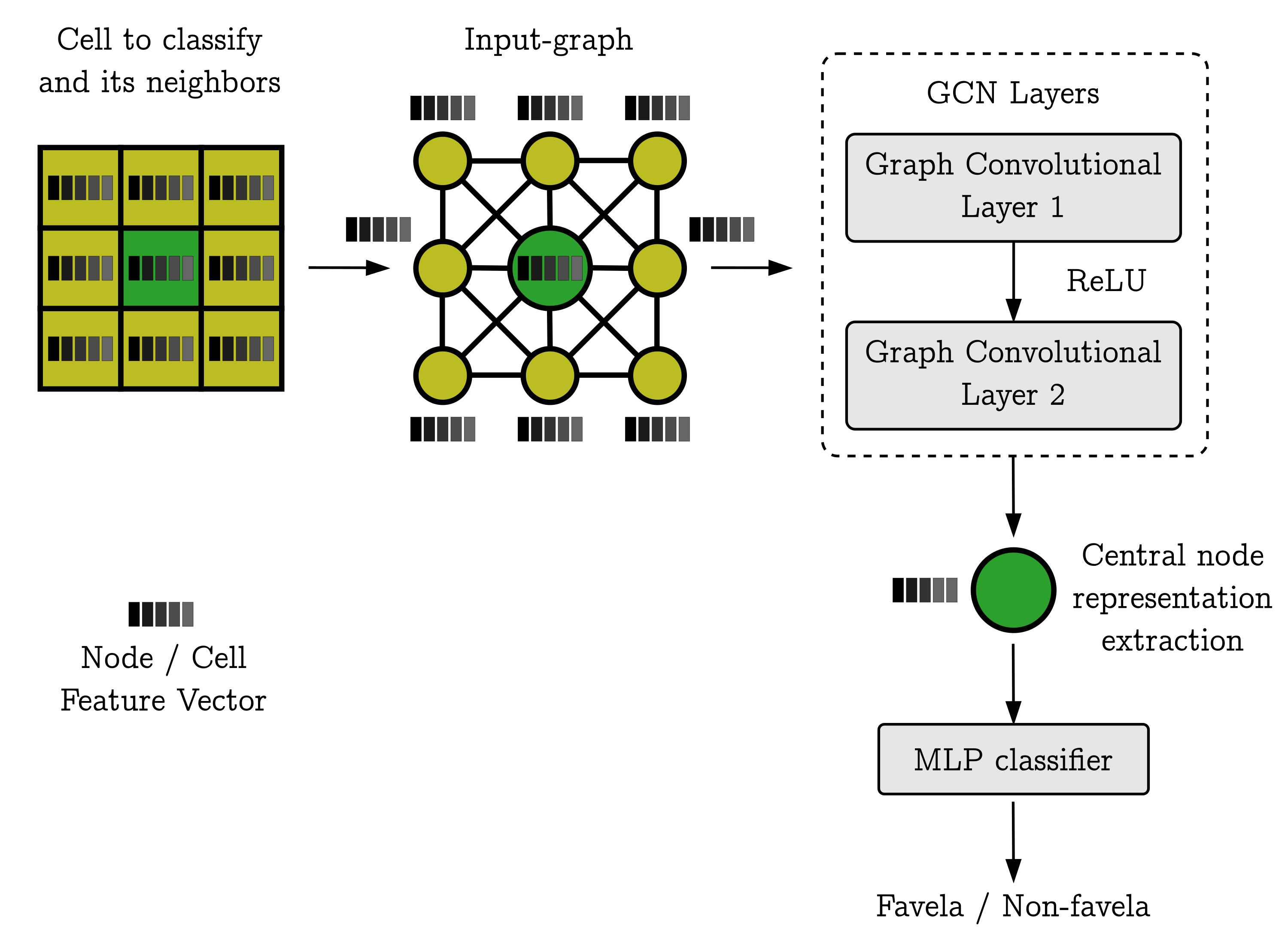}
\caption{Our Graph-Based Classification Approach} 
\label{fig:GCN}
\end{figure}

Our classification method, presented in Fig.~\ref{fig:GCN}, leverages graph-based learning at the local neighborhood level. 

\subsection{Feature Representation of Spatial Units (Cells)} 

The study area is divided into cells according to a regular 150$m$ resolution orthogonal grid (see Fig.~\ref{fig:GTG}). Each cell, or spatial unit, is characterized by a set of 9 handcrafted features presented in Table~\ref{tab:feature_summary}. These features are inspired by Owen and Wong~\cite{owenApproachDifferentiateInformal2013} and are described in detail in our previous work~\cite{hallopeauAddressingDataImbalance2025}, together with our precise labeling process and the motivations behind these choices. The features include variables derived from the Copernicus DEM, the OpenStreetMap road network, and a Sentinel-2 satellite image. Each cell is labeled as either \underline{favela} or \underline{urban non-favela}, based on whether more than 90\% of its area is covered by the reference favela polygon from the Brazilian Institute of Geography and Statistics (see~\cite{hallopeauAddressingDataImbalance2025}).
\begin{table}[t]
\centering
\caption{Summary of the Grid Features.}
\begin{tabular}{p{3.5cm}p{3.5cm}p{5cm}}
\toprule
\textbf{Data source} & \textbf{Feature name} & \textbf{Description} \\
\midrule
Sentinel-2 \newline satellite imagery 
  & \cellcolor{gray!2.5}Vegetation proportion & \cellcolor{gray!2.5}Proportion of surface covered by vegetation, calculated using NDVI threshold ($\geq 0.6$). \\
  & \cellcolor{gray!10}Entropy & \cellcolor{gray!10}Global entropy computed as the mean entropy across the 12 Sentinel-2 bands, normalized between 0 and 1. \\
\midrule
Copernicus \newline elevation model 
  & \cellcolor{gray!2.5}Slope & \cellcolor{gray!2.5}Average slope of all pixels in a cell, derived from elevation gradients. \\
  & \cellcolor{gray!10}Profile convexity & \cellcolor{gray!10}Average profile convexity of pixels, representing terrain curvature. \\
\midrule
OpenStreetMap \newline street network 
  & \cellcolor{gray!2.5}Number of street nodes & \cellcolor{gray!2.5}Total count of street intersection nodes within each cell. \\
  & \cellcolor{gray!10}Total street length & \cellcolor{gray!10}Total length of streets within each cell. \\
  & \cellcolor{gray!2.5}Node connectivity statistics & \cellcolor{gray!2.5}Mean, minimum, and maximum number of streets connected to each node within the cell. \\
\bottomrule
\end{tabular}
\label{tab:feature_summary}
\end{table}

\subsection{Graph Representation of Cell Neighborhood} 

For each labeled cell, we construct a local undirected graph capturing its immediate spatial neighborhood (see Fig.~\ref{fig:GTG}). The graph is composed of the target cell and its adjacent cells within a 3×3 window, corresponding to all neighboring cells that share an edge or a corner. Edges are defined based on spatial adjacency between cells, forming a planar neighborhood structure that reflects local topology and capture spatial relationships. This localized graph is used as input to the model, which predicts the label of the central node only.


\subsection{Classification with a Graph Convolutional Network} 
\label{subsec:GCN}
We introduce a lightweight two-layer Graph Convolutional Network (GCN). The graph is processed and only the representation of the central node is used for classification. This setup encourages the network to focus on the context around the central cell. The model is trained using the cross-entropy loss and evaluated based on the label of the central node only. The classification process proceeds as follows:
\begin{itemize}
    \item The graph is passed through a first GCN layer that aggregates features from neighboring nodes. The output is passed through a ReLU activation function, and a second GCN layer refines the representation using a new round of feature aggregation. A ReLU activation is again applied. The node features are successively mapped from $\mathbb{R}^9$ to $\mathbb{R}^{64}$, then from $\mathbb{R}^{64}$ to $\mathbb{R}^{64}$.
    \item The final representation of the central node (corresponding to the target cell) is fed into a fully connected layer with two output units. A softmax function produces the final class probabilities (informal versus formal).
\end{itemize}

\section{Experiments}

\subsection{Dataset}
\label{subsec:SA}

\subsubsection{Study Area}

Rio de Janeiro, located on the southeastern coast of Brazil, is a city marked by stark socio-spatial contrasts. Its dense urban fabric spans coastal plains, forested mountains, and steep hills, with affluent areas like Copacabana and Ipanema coexisting alongside informal settlements, called favelas. These contrasts reflect both colonial legacies and rapid 20th-century urban growth.

Favelas first emerged in the 1890s and have since multiplied across the city, initially in central areas, later expanding along transport corridors and into peripheral and hillside zones. Built primarily from exposed bricks with ceramic or concrete roofs, they are more permanent than many informal settlements elsewhere. The spatial distribution of favelas reflects complex historical, topographical, and socio-economic dynamics.

From a machine learning perspective, informal settlement classification in Rio de Janeiro represents a highly challenging testbed. Informal areas exhibit significant intra-class variability (e.g., morphology, density, materials), while some formal neighborhoods can resemble favelas, leading to low inter-class separability.

\subsubsection{Spatial Cross-Validation Zoning}
\label{subsubsec:SCV}

\begin{figure}[t]
\centering
\includegraphics[width=\textwidth]{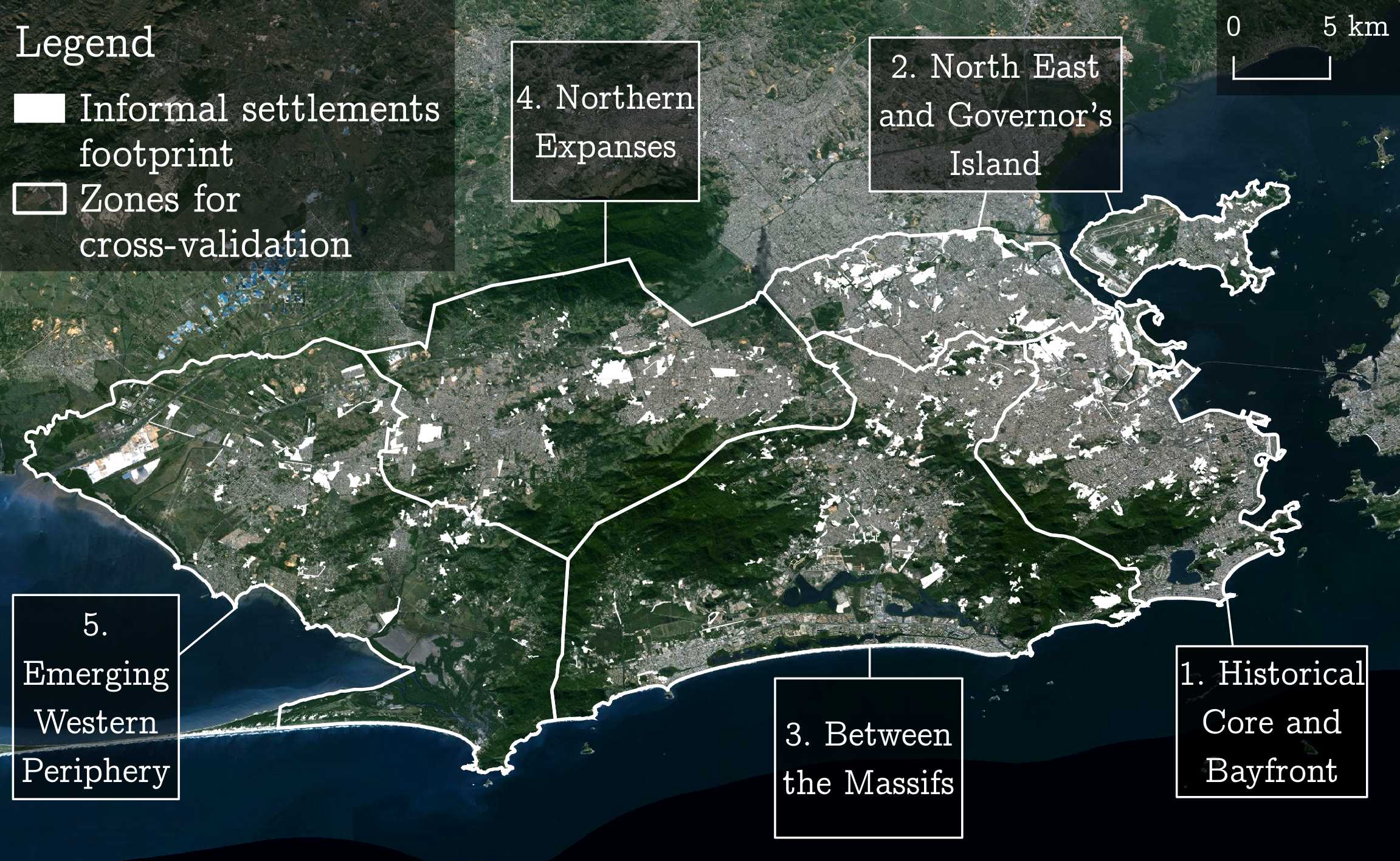}
\caption{Study Area in Rio de Janeiro Divided into Five Distinct Urban Zones used for Cross-Validation} 
\label{fig:SA}
\end{figure}

To validate our method given the spatial heterogeneity of the city, we partition our dataset into five custom zones for spatial cross-validation~\cite{robertsCrossvalidationStrategiesData2017}, as presented in Fig.~\ref{fig:SA}. These zones respect geographic and historical patterns while balancing favela distribution:
\begin{enumerate}
    \item Historical Core and Bayfront: Central coastal district bordered by the Tijuca massif, including early favelas like Jacarezinho and Complexo da Maré.
    \item North East and Governor’s Island: Industrial and residential zone hosting major favelas such as Morro do Dendê and Acari.
    \item Between the Massifs: Mountainous corridor with slope-based favelas like Rocinha and Vidigal.
    \item Northern Expanses: Zone north of Pedra Branca with both consolidated (e.g., Vila do Vintém) and emerging favelas.
    \item Emerging Western Periphery: Semi-rural frontier of urban growth with newer, smaller settlements like Cajueiro.
\end{enumerate}
Our zoning supports a robust evaluation of generalization across urban contexts with varying degrees of development. At each iteration, one region is held out for testing while the remaining four are used for training. To ensure spatial independence, no favelas or areas are shared between training and test sets~\cite{beigaiteSpatialCrossValidationGlobally2022}, and the model is re-initialized from scratch for each fold.

\subsubsection{Data Balancing} 

When designing machine learning models, balancing datasets is essential to prevent the error calculation from being skewed towards the majority class. In urban environments, cells corresponding to informal settlements are frequently under-represented compared to formal residential areas. In our dataset, there are approximately 30 times fewer favela cells than non-favela cells. To mitigate the class imbalance, we apply random undersampling~\cite{heLearningImbalancedData2009}. For each training or test set, an equal number of non-favela and favela cells is randomly selected.



\subsection{Baselines}

We design two baseline approaches based on a Multi-Layer Perceptron (MLP) classifier that operates on handcrafted features computed for each cell of the mesh. The two variants differs in the extent of spatial context provided to the model.
\begin{itemize}
\item \underline{MLP no-neighbor baseline.} This baseline uses only the 9-dimensional feature vector of the target cell as input. The spatial context provided by neighboring cells is ignored.
\item \underline{MLP neighbor baseline.} This variant incorporates spatial context by including the features of the 8 neighboring cells (i.e., those sharing an edge or a corner with the target cell), resulting in a input vector with $9 \times (1 + 8) = 81$ dimensions. If fewer than 8 neighbors are available, missing values are padded with zeros.
\end{itemize}

Comparing the first baseline to the second evaluates the benefit of using spatial context, while comparing the second baseline to our graph-based approach assesses whether simple feature concatenation is sufficient or if modeling spatial structure explicitly (as in our GCN) provides additional value. To ensure a fair comparison, all models use a comparable number of parameters: the MLP baselines use a single hidden layer of size 64, with input sizes of 9 or 81 depending on whether neighbor features are included, followed by a ReLU activation and an output layer of size 2. Our GCN model applies two GCN layers (9→64→64) followed by a MLP classifier (64→2) on the central node representation, maintaining similar model capacity while explicitly leveraging graph structure (see Section~\ref{subsec:GCN}). Training is conducted using the Adam optimizer and cross-entropy loss for $400$ epochs, with a batch size of $32$ and a learning rate of $0.001$. All experiments are run on an NVIDIA RTX 4000 Ada Generation GPU using PyTorch 2.4.1 and PyTorch Geometric 2.6.1.


\subsection{Metrics}

We evaluate model performance using 4 standard binary classification metrics: Precision, Recall, F1-score, and Kappa coefficient. Precision quantifies the proportion of predicted favela cells that are correct, while Recall captures the proportion of actual favela cells correctly identified. The F1-score balances these two metrics. The Kappa coefficient measures the agreement between predictions and ground truth, adjusting for chance. Together, these metrics provide a comprehensive assessment of detection quality.

To capture variability in performance, we repeat the entire evaluation protocol (see Section~\ref{subsubsec:SCV}) 10 times, yielding 50 metric values per evaluation (10 per region). We report the average and standard deviation of each metric per region. Finally, we compute global averages and standard deviations over the five regions to reflect inter-regional performance variability.

\section{Results and Discussion}




Table~\ref{tab:results_by_zone} presents average Kappa scores and standard deviations across 10 independent cross-validations for each zone and overall. Full results for other metrics are provided in the supplementary file tables.pdf, available on the project repository at \href{https://github.com/Hallopeau/gcn-informal-settlements}{gcn-informal-settlements}. All models perform unevenly across zones, with lower scores in heterogeneous areas such as zones 2 and 3. The baseline MLP model using only local cell features performs the worst. Incorporating neighboring cell features through simple concatenation significantly improves performance in all zones, confirming the value of local spatial context. Our GCN model further outperforms both baselines consistently across all regions, yielding a global Kappa gain of 10 points over the best MLP variant. 

These improvements highlight the value of structured message passing over naive concatenation. While neighbor-feature concatenation already provides substantial gains, our graph-based approach better captures spatial relationships and generalizes more effectively. This is true even in zones where local features suffice (e.g., zones 1 and 5), suggesting that spatial structure remains informative in coherent areas. Nonetheless, remaining variability across zones points to persistent challenges related to fragmented or irregular urban morphology.

\begin{table}[t]
\centering
\setlength{\tabcolsep}{7pt} 
\caption{
{Kappa Scores (\%) for the Baselines and our Approach Across the 5 Cross-Validation Zones and Overall.}
Best results are underlined.}
\begin{tabular}{lcccccc}
\toprule
Approach & Zone 1 & Zone 2 & Zone 3 & Zone 4 & Zone 5 & Global \\
\midrule
No-neighbors & $75 \pm 3$ & $43 \pm 10$ & $43 \pm 8$ & $61 \pm 3$ & $75 \pm 4$ & $59 \pm 14$ \\
Neighbors MLP & $74 \pm 4$ & $55 \pm 6$ & $56 \pm 9$ & $70 \pm 4$ & $76 \pm 3$ & $66 \pm 9$ \\
Our GCN & $\underline{86} \pm 3$ & $\underline{69} \pm 9$ & $\underline{66} \pm 4$ & $\underline{74} \pm 5$ & $\underline{87} \pm 2$ & $\underline{76} \pm 9$ \\
\bottomrule
\end{tabular}
\label{tab:results_by_zone}
\end{table}

\section{Conclusion}

In this work, we addressed the problem of classifying urban territories from spatial data including satellite imagery, with a specific focus on identifying favelas in Rio de Janeiro. By modeling spatial context explicitly through a graph-based representation, we proposed a novel framework that goes beyond patch-level classification and incorporates relationships between spatial units.

Our experiments demonstrate that incorporating neighborhood information substantially improves classification performance. While a simple concatenation of neighboring features already yields noticeable gains over local-only models, our graph convolutional approach further enhances generalization, particularly in the most heterogeneous regions of the city.

These results highlight the value of spatially structured learning in urban remote sensing. Beyond the specific case of favelas, our findings suggest that graph-based models can provide more robust and context-aware classification in urban environments characterized by strong spatial dependencies. Future work may explore richer graph construction strategies (e.g., learning the graph), incorporation of temporal dynamics, or multi-resolution fusion and other model architectures with attention mechanisms to further improve classification performance and spatial transferability. All code associated with this work is available on the project repository with additional results: \href{https://github.com/Hallopeau/gcn-informal-settlements}{\small gcn-informal-settlements}.

    
\begin{credits}
\subsubsection{\ackname} This study was funded by by the French Space Agency (CNES) through the CNES-TOSCA committee (tele-epidemiology thematic section).


\end{credits}
%
%
%
%

\bibliographystyle{splncs04}
\bibliography{bib}





\end{document}